\documentclass{ifacconf}
\usepackage{mathtools}
\usepackage{graphicx}      
\usepackage{natbib}        
\usepackage{algorithm}
\usepackage{algpseudocode}
\usepackage{multirow}
\usepackage{amsmath}
\usepackage{svg}
\usepackage{pdfpages}
\usepackage{amsfonts}

\DeclareMathOperator*{\argmin}{arg\,min}

\begin{document}
\begin{frontmatter}

\title{Safe Reinforcement Learning for an Energy-Efficient Driver Assistance System} 


\author[First]{Habtamu Hailemichael} 
\author[First]{Beshah Ayalew} 
\author[First]{Lindsey Kerbel}
\author[Second]{Andrej Ivanco}
\author[Second]{Keith Loiselle}

\address[First]{Automotive Engineering, Clemson University, Greenville, SC 29607, USA (hhailem, beshah, lsutto2)@clemson.edu.}
\address[Second]{Allison Transmission Inc., One Allison Way, Indianapolis, IN, 46222, USA (andrej.ivanco, keith.loiselle)@allisontransmission.com}

\begin{abstract}                
Reinforcement learning (RL)-based driver assistance systems seek to improve fuel consumption via continual improvement of powertrain control actions considering experiential data from the field. However, the need to explore diverse experiences in order to learn optimal policies often limits the application of RL techniques in safety-critical systems like vehicle control. In this paper, an exponential control barrier function (ECBF) is derived and utilized to filter unsafe actions proposed by an RL-based driver assistance system. The RL agent freely explores and optimizes the performance objectives while unsafe actions are projected to the closest actions in the safe domain. The reward is structured so that driver's acceleration requests are met in a manner that boosts fuel economy and doesn't compromise comfort. The optimal gear and traction torque control actions that maximize the cumulative reward are computed via the Maximum a Posteriori Policy Optimization (MPO) algorithm configured for a hybrid action space. The proposed safe-RL scheme is trained and evaluated in car following scenarios where it is shown that it effectively avoids collision both during training and evaluation while delivering on the expected fuel economy improvements for the driver assistance system. 
\end{abstract}

\begin{keyword}
RL driver-assist, Safe reinforcement learning, Safety filtering, Control barrier functions
\end{keyword}

\end{frontmatter}

\section{Introduction}
Reliable, safe, and efficient commercial vehicles are essential for the transportation industry to have a positive impact on the environment, the economy, and road safety. Given the estimated increase in freight demand of $16\%$ by 2030~\citep{faf}, there is clearly a need and an opportunity to reduce emission and fuel usage as more of these vehicles get on the roads to meet this demand. Furthermore, ensuring safety via accident prevention is critical. Advanced driver assistance systems (ADAS) such as emergency braking, adaptive cruise control (ACC), and lane keeping assist have been developed to primarily address the safety concerns. More advanced systems additionally aid the driver with ecological (fuel saving) driving behaviors such as reduced braking and accelerations~\citep{ecodriv}, optimizing velocity profiles for ACC~\citep{MaamDP,NieACCMPC} and optimizing gear shifting ~\citep{optgear}.

As typical commercial routes include frequent stopping and starting along with various required speeds, the ACC approach may be cumbersome for a driver to use. In~\cite{Yoon2020}, a driver assistance implementation is proposed that uses radar information and motion models to directly modulate the torque request to the powertrain/braking system. To this end, an MPC scheme is employed to optimize traction/braking torque/power while tracking the driver's desired acceleration and maintaining a safe distance to a leading vehicle. In~\cite{kerbel}, a similar driver assistance objective is pursued in a model-free reinforcement learning approach to learn both optimal gear selection and torque request using fuel usage and other reward signals in the vehicle's experience. Although this study demonstrated a fuel consumption improvement of up to $12\%$, it did not include provisions to guarantee collision avoidance. Unlike typical optimal control schemes such as MPC, where hard constraints are set based on a dynamic model, it is generally difficult to enforce such constraints in RL controllers where learning an optimal policy requires exploration of different actions and states. However, unlimited exploration is unacceptable for safety-critical systems such as vehicle control. In this paper, we construct a driver-assist RL agent that targets fuel efficiency and driver accommodation and incorporates elements that ensure safety.

Different approaches are proposed to properly constrain the exploration of the RL agent within a safe set. In ~\cite{StepGear}, a supervisor is used to simply enforce (override) the gear and engine speed constraints to control the transmission, yet the RL agent never really learns these limits. Often, reward shaping approaches are utilized by assigning a penalty to safety violations that discourage policies leading to constraint violation. Since the RL agent with reward shaping learns the penalties only after experiencing them, this approach does not guarantee safety, especially during initial training. Another approach to enforcing safety is to pose the problem as a constrained Markov decision process (CMDP) where a constraint cost is assigned for each state-action pair and the RL agent learns to keep the discounted constraint cost over the horizon below a certain threshold \citep{Altman}. Many implementations of the CMDP then involve joint optimizations of the main performance task and the constraint reward, and this entails trade-offs between safety and performance. In this work, we seek to somewhat decouple the two goals by adopting what is known as a safety filtering approach. This approach configures the RL agent to focus on maximizing performance (reward), while a safety layer/filter is designed to project the outputs of the RL agent onto a safe set. Although the filter does not typically interfere with the inner workings of the RL agent, it will influence performance as it often determines the extent of the safe set and subsequent interactions of the RL agent with the system under control. Evaluations of the proposed actions in the safety layer could be based on learning constraints \citep{Dalal2018} and safety indexes \citep{Thananjeyan2021,Srinivasan2020} from offline data or using a dynamic model of the system.

Of the dynamic model-based approaches to safety filtering, control barrier functions (CBF) provide scalable and computationally light safety filters \citep{li2021comparison}. A CBF applies hard safety constraints by forcing the system to operate in the invariant safe-set defined by a super-level set of a continuously differentiable function $h(x):R^{n}\rightarrow R$. The actions selected by the RL agent are projected into the safe set in such a manner that the proposed actions are minimally modified ~\citep{Ames2019}, and no unsafe actions are passed to the controlled system. One could come up with handcrafted CBFs considering the dynamics of the system; a case in point is the relationship between the maximum deceleration available to the vehicle and the distance gap in the collision avoidance problems \citep{Ames2014,Cheng2019}. For high relative degree nonlinear systems, as in the present application, tailored CBFs known as exponential barrier functions (ECBF) have been proposed \citep{Nguyen2016}.

In this paper, we derive a specific ECBF structure that works in conjunction with the RL driver-assist agent in order to take explicit consideration of inertia effects which are relevant for the safety of commercial vehicles in traffic.
The main performance goal of the driver-assist RL agent is given by a multi-objective reward function that is structured to balance driver accommodation, fuel economy, and smooth vehicle operation. In addition, driveability is encouraged by introducing an additional reward for reserve power. In this regard, \cite{Ngo2012} characterizes acceleration potential at a given speed by merely analyzing different standard drive cycles. In this paper, we propose to learn the power reserve reward to customize the vehicle's response to the driving conditions and the driver's tendencies.

To summarise, the contributions of this paper are: $1)$ formulation of a driver assist RL agent configured for reward optimal gear selection and torque control of a commercial vehicle, $2)$ derivation of an ECBF safety filter to work with this RL agent and $3)$ demonstration of the potential of learning power reserve attributes to further customize the system to the driver and driving conditions. The rest of the paper is organized as follows: Section \ref{sec: II} discusses the vehicle model and the driver-assist RL agent. Section \ref{sec: III} discusses the design of the safety filter and the subsequent projection of the output of the RL agent onto the safe set. Section \ref{sec: IV} presents simulation and training settings and results are discussed in Section \ref{sec: V}. Finally, Section \ref{sec: VI} concludes the paper.

\section{Vehicle Environment and RL controller} \label{sec: II}
\begin{figure}
\begin{center}
\includegraphics[width=8.4cm, height= 2.2 cm]{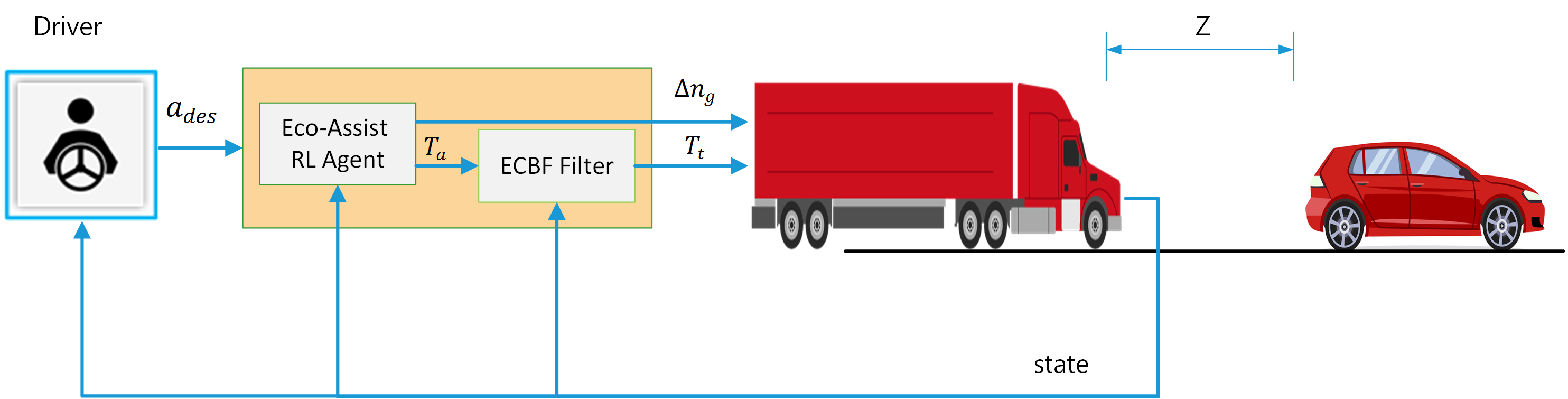}  
\caption{Proposed safe RL-based Eco-Assist system set up} 
\label{fig:DAsetup}
\end{center}
\end{figure}

A schematic of the proposed RL-based driver assistance system, including the safety filter, is shown in Fig.\ref{fig:DAsetup}. In this section, we detail the different computational components of the Driver-Assist RL agent; the next section deals with the ECBF safety filter.

The vehicle-driver-environment is modeled as Markov decision process (MDP) with state $s$, actions $a$, rewards $r$ and a discount factor $\gamma$. The states are included in $s=\{v_l, v_{rel}, a_{des}, a, z, n_{g}, m_v, \theta, f\}$ which, respectively, are the ego vehicle velocity, the relative velocity between the preceding and ego vehicle, the driver demanded acceleration, the actual vehicle acceleration, the separation distance with the preceding vehicle, transmission gear, mass of the vehicle, road grade and a flag to alert if a preceding vehicle is the sensing range of the ego-vehicle's radar. The RL controller is designed to maximize the vehicle's performance objectives through wheel traction torque $T_t$ control and gear change selection $\Delta{n_g}$, i.e., the action vector is: $a =\{T_t,\Delta{n_g}\}$. The velocity and the wheel traction torque are propagated back to calculate the engine torque and speed using the transmission ratio of the current gear and the final drive ratio. A fuel rate map is then utilized to solve for the fuel consumption at the given engine torque and speed.

The reward function, given by (\ref{eq:outofrange}) below, is structured to capture the performance objectives of the driver-assist RL-agent. The major objective of the RL agent is to fulfill the driver's acceleration request, and consequently, an acceleration error term is given a higher weight, $w_a$. Through the fuel rate reward term, weighted by $w_{f}$, the RL agent is encouraged to operate the engine at fuel-efficient operating points while fulfilling the driver-demanded acceleration. Smooth torque changes are weighted with $w_t$, and gear hunting and the associated rough vehicle operation are mitigated by including a shifting frequency penalty term weighted by $w_g$. Note relevant reward signals are normalized by their corresponding maximum values as noted by the max subscripts. $\dot{m}_f$ and $\Delta{T_{t}}$ are the fuel rate and torque change respectively.

\begin{equation} \label{eq:outofrange}
\begin{split}
r=w_a0.1^\frac{{|a}-a_{des}|}{a_{des,max}}+w_{f}0.1^\frac{{\dot{m}}_f}{m_{f,max}} + w_{T} 0.1^\frac{{|\Delta T_t}|}{T_{t,max}} +\\ w_{g}0.1{^\frac{{|\Delta n_{g}}|}{n_{g,max}}} + r_{pr},
\end{split}
\end{equation}
 where $r_{pr}$ models the power reserve reward term that accounts for enhanced driveability. We define it as: $r_{pr}= w_{pr}0.1^\frac{{P_{res,req-P_{res}}}}{P_{res,req}}$ if $P_{res} < P_{res,req}$, else $r_{pr}=w_{pr}$, where $w_{pr}$ is the corresponding weight. $P_{res}$ is the actual available power which is given in terms of engine
speed and engine torque as $P_{res} = (T_{e,max}(\omega)-T_e)\omega$; and $P_{res,req}$ is the required power reserve which we discuss next.
 
 To adapt the $P_{res,req}$ with the different acceleration demands in different driving conditions, \cite{Ngo2012} models acceleration potential as varying with the vehicle velocity. To this end, the speed of the vehicle is discretized and acceleration requests for each speed level in multiple standard cycles are collected to be fitted in a cumulative probability distribution. The maximum acceleration at a given design confidence level (usually $90 \%$) is taken as the required acceleration potential at that velocity, $a_{req}(v)$. Given $a_{req}(v)$, the required power reserve is then modeled with $P_{res,req} = m_vva_{req}(v)$. In our work, rather than using acceleration data from standard drive cycles, we propose using the data generated by the driver in the prevailing driving conditions. The demanded acceleration of the driver is continuously fitted to get $a_{req}(v)$ that adapts to the driver's demand. Observing that polynomial fits of $a_{req}$ suggested in \cite{Ngo2012} lead to overfitting issues when used with driver generated training data, we instead use a logistic function that is easier to parametrize and learn:
\begin{equation} \label{eq:logisticfunction}
\begin{split}
{{
a_{req}(v)=\frac{k_1}{1+k_2k_3^{-v}}.
}}
\end{split}
\end{equation}

Next, we briefly describe the framework we adopted for training the driver assist RL agent. The states, control actions, next states, and associated rewards are continuously stored in the memory buffer $\mathcal{R}$. 
We use actor-critic architecture proposed by~\cite{kerbel} that utilizes the off-policy algorithm known as maximum posteriori optimization (MPO) \citep{abdolmaleki2018, Neunert2020} for training. Even if it is possible to use other state of the art algorithms, we use MPO for its sample efficiency and robustness to hyper-parameters as well as ease of use with the hybrid action space for the present problem. The algorithm starts with a policy evaluation step where a critic network approximates the state-action-value (Q-value) for the policy. A squared loss function is minimized between the current Q-value and an estimated target Q-value, $Q_{target}(s,a)$. For this study, we adopted the Retrace algorithm, known for efficiency and stability, as described in~\cite{ZhuRetrace} for our target Q-value.

For policy improvement (actor network), the MPO algorithm uses an expectation-maximization scheme. By taking samples from the memory buffer, we construct a non-parametric policy $q$ that maximizes $\mathbb{E}_q [Q_{\theta_k}(s,a)]$. 
\begin{equation} \label{MPOone}
\begin{split}
& \max_{q} \mathbb{E}_{q(a|s)} [Q_{\theta_k}(s,a)] \\
& s.t.\; \mathbb{E}_{\mu(s)} \left[ KL \left( q(a|s)|| \pi_{\theta_k}(a|s) \right) \right] < \epsilon,
\end{split}
\end{equation}
where $\mu(s)$ is the visitation distribution given in the replay buffer. Then a new parametric policy $\pi_{\theta}$ is fitted to $q$ with a Kullback–Leibler (KL) divergence constraint to limit excessive deviations from the current policy. The parameters of the actor network are updated via a gradient-based optimization in Adam solver~\citep{Kingma2015AdamAM}.
More detailed explanation of the MPO algorithm can be found in~\cite{Neunert2020} and ~\cite{abdolmaleki2018}. Further implementation details for the present application can also be found in our straight RL implementation in~\cite{kerbel}. As noted above, other state of the art RL training algorithms can also be applied for the driver-assist RL agent and this is independent of the safety filter discussed next.

\section{Exponential CBF Safety filter } \label{sec: III}
In this section, we give the derivation of the ECBF filter for our application. We start with a brief review of the definition of CBF and ECBF. We refer readers to \cite{Nguyen2016} for more detailed discussions on these topics.
 Consider a nonlinear control affine system:
\begin{equation} \label{eq:affinedynamics}
{{\dot{x}=f\left(x\right)+g\left(x\right)u,}}
\end{equation}
where $f$ and $g$ are locally Lipschitz, $ x\in\mathcal{R}^n $ is the state of the system, $ u\in\mathcal{R}^m $ is the control input. Assume a safe set defined by $\mathcal{C}=\left\{x\in\mathcal{R}^n|h\left(x\right)\geq0\right\}$, where $h:\mathcal{R}^n\rightarrow\mathcal{R}$ is a continuously differentiable function. Then $h$ is a CBF if there exists an extended class $ \kappa_\infty$ function $\alpha$ such that for all $x\in Int\left(\mathcal{C}\right)=\left\{x\in\mathcal{R}^n:h\left(x\right)>0\right\}:$
\begin{equation} \label{eq:CBF}
{{
\displaystyle\sup_{u\in U}{\left[L_fh\left(x\right)+L_gh\left(x\right)u\right]}\geq-\alpha\left(h\left(x\right)\right)
}}.
\end{equation}
The fact that $h$ is a CBF ensures the safe set $\mathcal{C}$ is forward invariant and we are able to guarantee safety. ECBFs use input-output (IO) linearization of nonlinear systems with relative degree $r$ in order to generate CBFs. As detailed in \cite{Nguyen2016}, the new virtual linear system (after IO linearization) has state variables $\eta_b := [h(x),\dot{h}(x), \cdot \: \cdot \: \cdot,h^r (x)]^T$, input $ \mu$ and output $ h\left(x\right):$ 
\begin{equation} \label{eq:statefeedbacklinearizaion}
 \begin{aligned}
\dot\eta_{b} &= F\eta_{b}\left(x\right)+G\mu,\\
h(x) &= C\eta_{b},
\end{aligned}
\end{equation}
where $F$ and $G$ are matrices representing an integrator chain, $C=[1,\:0,\: \cdot \: \cdot\: \cdot \:,0]$.

The control action for the virtual linear system $\mu$ is the $r^{th}$ derivative of the control output; $\mu=L_f^{r}h(x) + L_gL_f^{r-1}h(x)u$. When $\mu$ is set by state feedback control with gain ${K}_\alpha$ as $\mu={-K}_\alpha\eta_b$, the control output evolves with time as $h\left(x\left(t\right)\right)=Ce^{\left(F-GK_\alpha\right)t}\eta_b\left(x_0\right)$. For initial condition $h\left(x_0\right)>0$, by imposing $\mu>{-K}_\alpha\eta_b\left(x\right)$, it possible to guarantee $h(x(t))\geq Ce^{(F-GK_\alpha\ )t}\eta_b\ (x_0\ )$. This relationship leads to the definition of ECBF. Considering the dynamic system (\ref{eq:affinedynamics}) and the set $\mathcal{C}=\left\{x\in\mathcal{R}^n\right|h\left(x\right)\geq{0}\}$, $h\left(x\right)$ is an ECBF if there exists $K_\alpha\in\mathcal{R}^{r\times1}$
\begin{equation} \label{eq:ecbf}
 \begin{aligned}
sup{\left[L_f^rh\left(x\right)+L_gL_f^{r-1}h\left(x\right)u\right]}\geq{-K}_\alpha\eta_b\left(x\right),\ \forall\ x\in\mathcal{C}.
\end{aligned}
\end{equation}

If $K_\alpha$ makes the closed-loop system matrix $F-GK_\alpha$ stronger than Hurwitz and for favorable initial conditions choosing, $\mu\geq-K_\alpha\eta_b\left(x\right)$ guarantees $h\left(x\right)$ is an ECBF. Pole placement strategies of linear feedback control can then be used to design the ECBF.

We construct the collision avoidance model with the separation distance $z$, the velocity of the ego vehicle $v_h$ and velocity of the leading vehicle $v_l$ as state variables. The model follows:
\begin{subequations}\label{eq:ACCsetup}
 \begin{equation}
    \label{eq:ACCsetup-a}
      {{\dot{z}=v_l-v_h}},
  \end{equation}
  \begin{equation}
    \label{eq:ACCsetup-b}
    {{\dot{v_l}=a_l}},
  \end{equation}
  \begin{equation}
    \label{eq:ACCsetup-c}
    {{{\dot{v}}_h=\frac{T_t}{r_wm_v}-\frac{F_r\left(v_h,m_v,\theta\right)}{m_v}}},
  \end{equation}
\end{subequations}

\begin{equation} \label{eq:resistanceforces}
{{
F_r=\frac{\rho A c_dv_h^2}{2}+m_vgf\cos{\theta}+m_vg\sin{\theta},
}}
\end{equation}
where $F_r$ is the total resistance force that includes gravitational, rolling friction and aerodynamics resistances. $T_t, c_d, f, \theta,m_v, \rho, A_v, r_w, a_l $ are the traction torque at the wheels, aerodynamic coefficient, rolling resistance coefficient, road grade, the mass of the vehicle, the density of air, the frontal area of the vehicle, the radius of the wheels, and acceleration of the leading vehicle, respectively.

Having affine dynamics, the above state space representation could also be separated into unactuated dynamics $f\left(x\right)$ and actuated dynamic $g\left(x\right)$ components when written as (\ref{eq:affinedynamics}). With the choice of a minimum inter-vehicle distance objective $z_0$, a natural choice is $h\left(x\right)=\ z-z_0$. Input-output linearization is then employed to transform the nonlinear dynamics into a virtual linear system as in (\ref{eq:statefeedbacklinearizaion}) and following the accompanying discussions above, with the feedback gain ${K}_\alpha = [{k}_{\alpha1}, {k}_{\alpha2}]$, we have:
\begin{equation} \label{eq:hx}
{{
\dot{h}(x) = v_l-v_h,
}}
\end{equation}
\begin{equation} \label{eq:hddot}
{{
\mu=\ddot{h}\left(x\right)=\frac{F_r\left(v_h,m_v,\theta\right)}{m_v}+a_l-\frac{T_t}{m_vr_w},
}}
\end{equation}

\begin{equation} \label{eq:kaeta}
\begin{aligned}
{-K}_\alpha\eta_b\left(x\right) &= -k_{\alpha1}h(x)-k_{\alpha2}\dot{h}(x) \\
 &=-k_{\alpha1}\left(z-z_0\right)\ -k_{\alpha2}\left(v_l{-v}_h\right).\\
\end{aligned}
\end{equation}

Using these with (\ref{eq:ecbf}), we arrive at the ECBF filter. The traction torque actions proposed by the RL agent are then passed through this safety filter before being sent to the vehicle environment. As shown in Fig. \ref{fig:DAsetup}, the ECBF filter enforces safety by projecting the action proposed by the RL agent $T_a(s)$ to the safe control traction torque $T_t$ in a way that introduces minimal change, as given by the QP problem below.
\begin{equation}
\begin{aligned}
 T^{*}_t = & \displaystyle\argmin_{T_{t}} \frac{1}{2}\left \| T_{t}-T_{a}(s) \right \|^2 \\
\textrm{s.t.} \quad & a_l+\frac{F_r\left(v_h,m_v,\theta\right)}{m_v}-\frac{T_t}{m_vr_w}\geq{-k}_{\alpha1}\left(z-z_0\right)\\
\quad & - k_{\alpha2}\left(v_l-v_h\right)\\
\end{aligned}
\end{equation}

\section{Simulation and Training Settings} \label{sec: IV}
Simulations of a medium-duty truck with a 10-speed automated manual transmission (AMT) are used to demonstrate the workings and performance of RL driver-assist agent with the ECBF safety filter. The actor and critic are represented with deep neural networks with three hidden layers, and each layer consists of 256 nodes. The RL controller is trained in a scenario in which the truck driven by an imperfect driver follows a preceding vehicle under Federal Test Procedure (FTP-75) drive cycle \citep{Barlow}. The agent is trained for a weight range of 5 to 10 tons, as commercial vehicles have to operate in significant load fluctuations. In order to help capture different driving experiences, the training data is randomized by adding noise to the velocity profile of the preceding vehicle, using random initial separation distance and road grade, and manipulating the parameters of the driver model, which in this work is taken as the intelligent driver model (IDM)\citep{TreIDM} described by the equations below. The desired minimum gap for IDM is calculated as in (\ref{eq:desiredz}), in which the approach term $z_{app}$ (\ref{eq:zapp}) dominates as the host vehicle approaches the preceding vehicle. 
By only adding $z_{app}$ for distance gaps closer than a certain threshold and ignoring this term beyond that, we emulate a distracted driver requesting unsafe actions.

\begin{figure}
\begin{center}
\includegraphics[width=8.4cm, height= 6 cm]{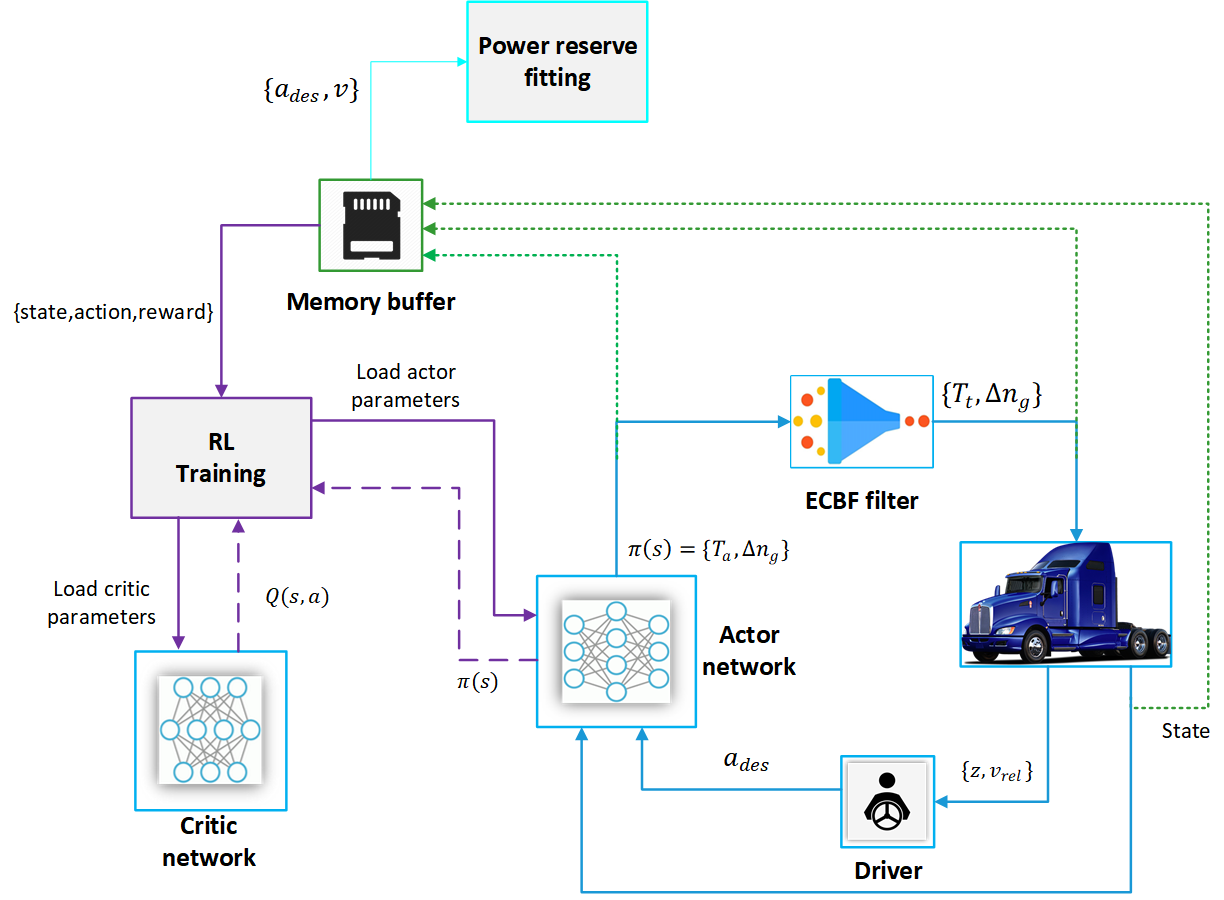}  
\caption{RL training setup} 
\label{fig:trainingsetup}
\end{center}
\end{figure}

\begin{equation} \label{eq:ades}
a=a_{max}\left(1-\left(\frac{v}{v_0}\right)^4-\left(\frac{z^\ast\left(v,v_{rel}\right)}{z}\right)^2\ \right),
\end{equation}
\begin{equation} \label{eq:desiredz}
z^\ast\left(v,v_{rel}\right)=z_0+\tau v-z_{app},
\end{equation}

\begin{equation} \label{eq:zapp}
z_{app}=\frac{v_{rel}v}{2\sqrt{a_{max}b}},
\end{equation}
where $a_{max}$ is the maximum acceleration, $b$ is a comfortable deceleration level, $\tau$ is the time headway.

In each simulation step, as shown in Fig. \ref{fig:trainingsetup}, the IDM driver requests acceleration based on the distance gap and velocity of the preceding vehicle. To fulfill the driver's demand, the actor network outputs $\{T_a,\Delta{n_g}\}$ and the proposed actions are filtered by the ECBF safety layer. The safe actions are then implemented in the vehicle environment and rewards are observed. As mentioned previously, the reward aims to fulfill the driver's acceleration demand in a manner that promotes fuel economy, driveability, and smooth vehicle operation. Accordingly, the reward objectives are weighted as $[w_a=0.65,w_f=0.2,w_T=0.05,w_g=0.05,w_{pr}=0.05]$. Concurrently, the acceleration demand of the driver and the vehicle velocity are used to characterize the power reserve. The maximum acceleration request of the driver (with $90\%$ confidence level) is fitted to a logistic function to continually adapt the power reserve term as described in Section \ref{sec: II}. 

For the vehicle described above and with the parameters given in Table 1, we designed an ECBF safety layer with the gain vector $K_{\alpha}=[0.8,2]$. We found that, for the given vehicle, the filter is effective in projecting unsafe actions with no collisions to report throughout the training. We observed that as the training progresses, the RL agent learns to control the vehicle's acceleration to align with the driver's request. The MPG is also improved with training, which shows the RL agent managed to learn to achieve the acceleration-tracking objective in a fuel-efficient manner. We omit the details of this training progression for space reasons, and instead present comparative evaluations in the next section.

The parameters we used for the simulation and training of the driver assist RL agent are given in Table \ref{tb:parametersMPO}.

\begin{figure}
\begin{center}
\label{fig:afit}
\end{center}
\end{figure}

\begin{table}[hb]
\begin{center}
\caption{Vehicle environment and RL hyperparameter setting}\label{tb:parametersMPO}
\begin{tabular}{cc|cc}
\hline
\multicolumn{2}{| c |}{Vehicle Parameters} & \multicolumn{2}{| c |}{MPO Hyperparameters}  \\\hline
Mass & 5\ - 10 tons & Actor, critic learning rate & ${10}^{-5},{10}^{-5}$ \\
$A_u$ & $7.71m^2$ & Dual constraint& 0.1 \\
$C_d$& $0.08$ & Retrace steps& 15 \\
$r_w$& $0.498$ & KL constraints $\epsilon_\mu,\epsilon_\sigma,\epsilon_d$ & $0.1,0.001,0.1$ \\
$f$& $0.015$ & $\alpha_d,\alpha_c$ & 10\\
$z_{sr}$&$350$ & $\gamma$ & 0.99 \\\hline
\end{tabular}
\end{center}
\end{table}

As a baseline, we also consider and simulate the same driving scenarios without the safe RL-assist agent in the loop (IDM only). The baseline powertrain control generates traction torque that compensates for resistances and fulfills the IDM driver's requested acceleration. For gear decisions, an optimal gear with the lowest fuel rate is selected according to a scheme described in \cite{Yoon2020}, which is model-based and has full knowledge of the engine fuel consumption map. Note that our RL agent has no such knowledge of the engine's fuel consumption map or any of the modeled dynamics.

\section{Evaluation RESULTS AND DISCUSSIONS} \label{sec: V}

\begin{figure}
\begin{center}
\includegraphics[width=8.6cm, height= 7.2 cm]{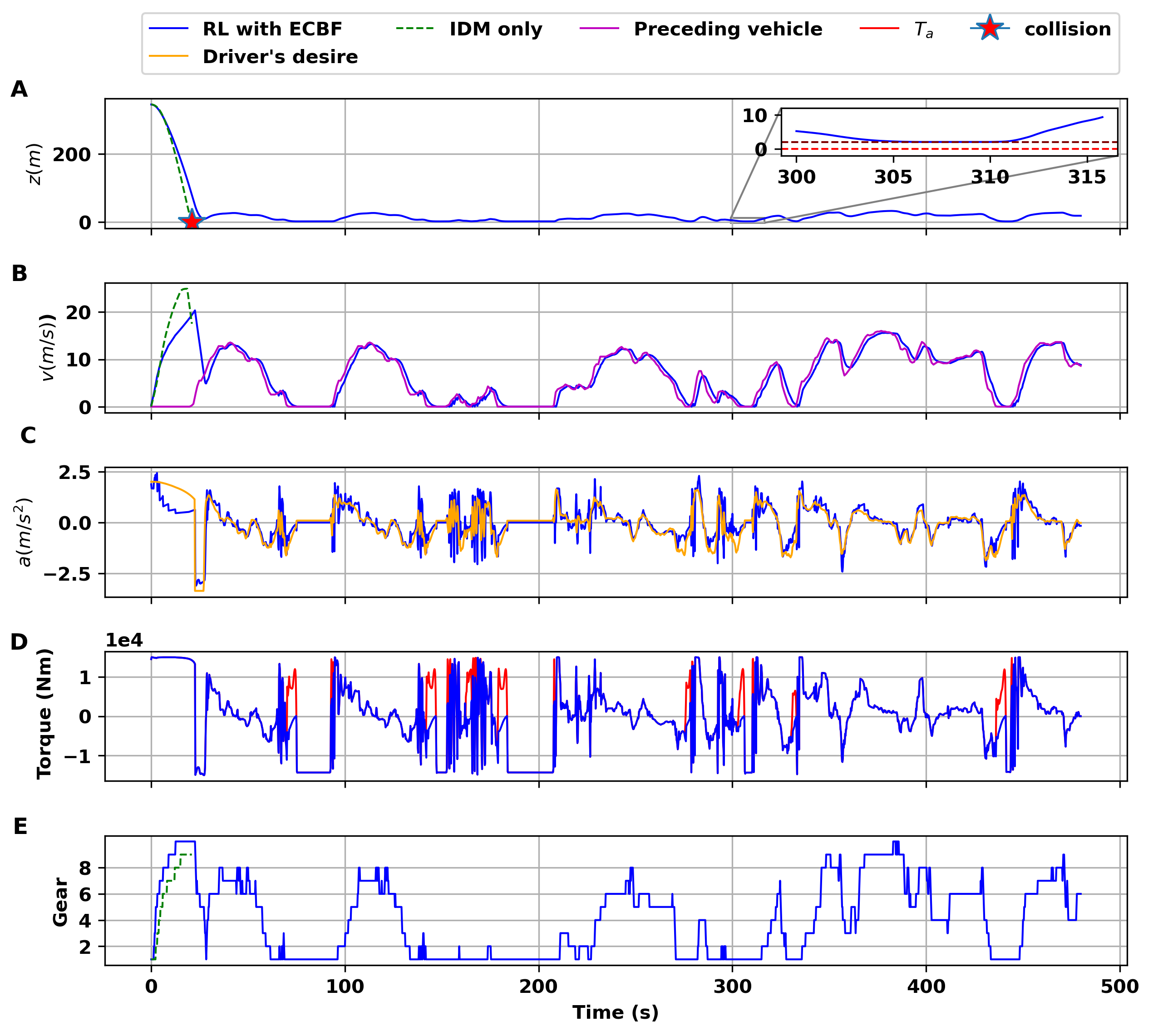} 
\vspace*{-3mm}
\caption{ Evaluation of RL assist with ECBF filter after a few training episodes for a shortsighted IDM driver. The red star in (A) shows collision for a case without the RL agent (IDM only).} 
\label{fig:Baddriver}
\end{center}
\end{figure}

The safety performance of the RL-ECBF assist is evaluated during and after training following the preceding vehicle under ARTEMIS Urban drive cycle \citep{Barlow}, which is different from the FTP cycle used for training. Fig. \ref{fig:Baddriver} illustrates how the RL-ECBF assist handles the worst case of training in which both the RL exploration and distracted driver are the sources of unsafe actions. The distracted driver is modeled by IDM with a $\tau$ of $2 seconds$ that only considers the approach term ($z_{app}$) for distance gaps less than $50 m$. Such a driver closes the initial $350 m$ separation and collides with the preceding vehicle (red star on Fig. \ref{fig:Baddriver}.A). However, when RL-ECBF assist is introduced, no unsafe action is sent to the vehicle environment due to significant ECBF projections (Fig. \ref{fig:Baddriver}.D) and the fact that the RL becomes aware of the safety boundaries. 

\begin{figure}
\begin{center}
\includegraphics[width=8.6 cm, height= 7.2 cm]{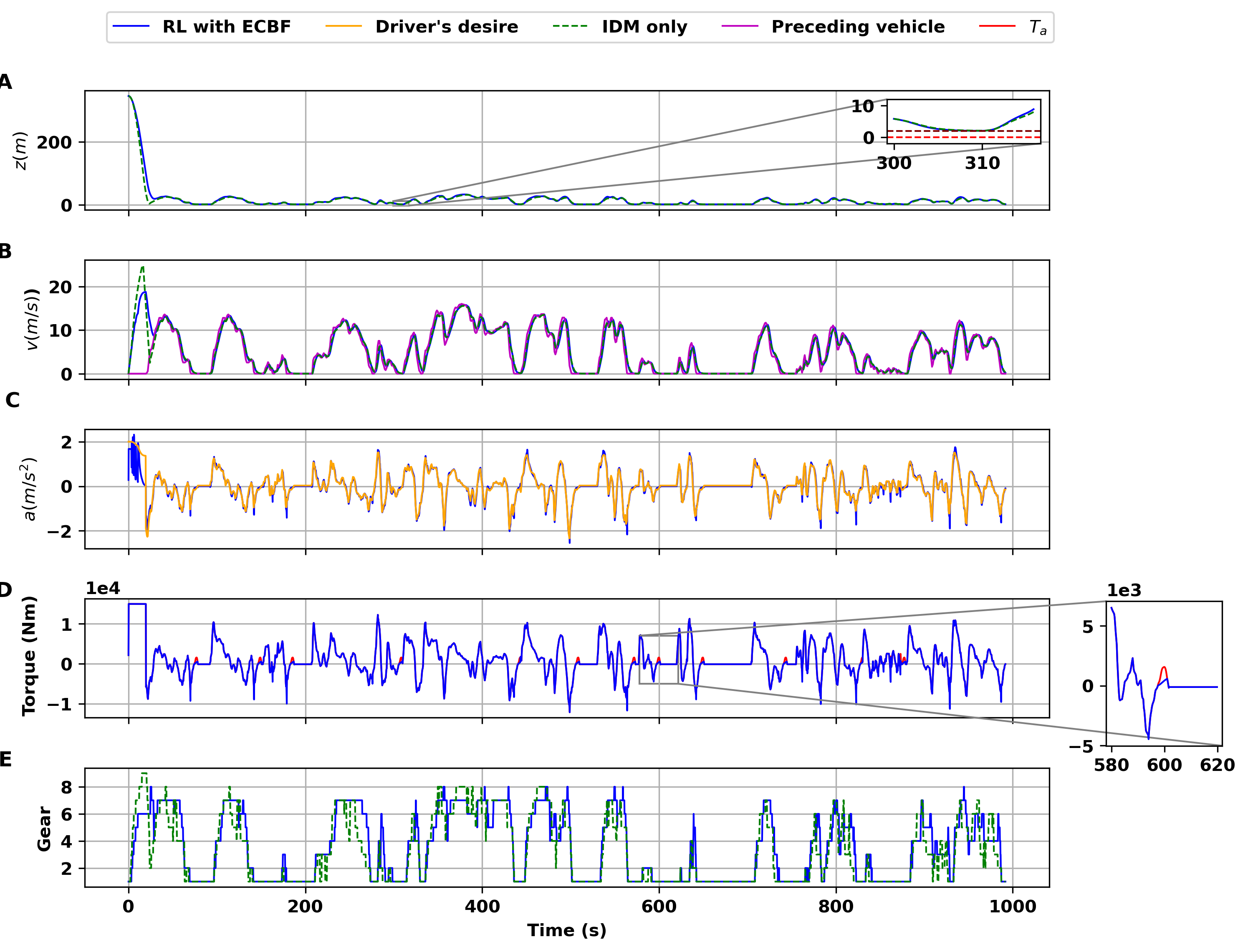} 
\vspace*{-3mm}
\caption{Evaluation of a well trained RL assist with ECBF filter for a conscientious driver} 
\label{fig:gooddriver}
\end{center}
\end{figure}

 We also looked at the performance of the RL-ECBF assist system with respect to meeting driver demand and improving fuel efficiency when paired with a conscientious driver. To evaluate this aspect, we model a relatively conscientious (good) driver via an IDM driver with the approach term ($z_{app}$) activated for distance gaps less than $100 m$. Fig.\ref{fig:gooddriver} and Table \ref{tb:Performance} show the performance comparison of the good IDM driver only case (with model-based gear and torque control) and when the same driver is assisted by a well-trained RL with ECBF safety filter. For the IDM driver only case, the root mean square error between the driver demand and the actual vehicle acceleration is $0.38$, and this value is improved to $0.17$ when the assist system is introduced. In addition to enhancing driveability, the RL agent tends to operate at higher gears, resulting in MPG improvement of $6.34\%$ over the baseline (IDM only with model-based powertrain control). The RL agent eventually learns to confine operations predominantly within the safe set defined by the ECBF filter, as the projection of unsafe actions usually produces suboptimal behavior (poor reward). This fact is illustrated in Fig. \ref{fig:gooddriver}.D in which the ECBF projections are relatively infrequent, small and limited to fast approaches in close proximities.

\begin{table}[hb]
\begin{center}
\caption{Performance comparison between conscious driver with and without RL-ECBF assist }\label{tb:Performance}
\begin{tabular}{p{1.25cm}|p{3cm}p{3cm}}
\hline
& IDM without \par RL-ECBF assist & IDM with RL-ECBF assist \\\hline
MPG & {\quad}$ 6.875 (-)$ &{\quad}{\quad} $7.31 (6.34\%)$ \\
$a_{rms}(m)$ & {\quad}{\quad} $0.38$ &{\quad}{\quad} $0.17$ \\ 
$Z_{mean}(m)$ &{\quad}{\quad} $15.5$ &{\quad}{\quad} $16.7$ \\ 
$Z_{min}(m)$ &{\quad}{\quad} $2$ &{\quad}{\quad} $2$ \\ \hline
\end{tabular}
\end{center}
\end{table}
\section{Conclusion} \label{sec: VI}
In this paper, a Driver-assist RL agent is formulated and demonstrated that can assist drivers in achieving better fuel economy and driveability. Safety is instilled into the RL agent by filtering unsafe actions using exponential control barrier functions (ECBF) both during training and actual operation. The RL-ECBF assist system is trained to maximize a multi-objective reward structure that balances the fulfillment of the driver’s acceleration demands, fuel economy, smooth operation and power reserve objectives. The acceleration request profile for a given driver is continuously adapted during the training of the RL agent to ensure enough acceleration potential is available for the particular driver. Evaluations on a different drive cycle than the agent is trained on demonstrated that the RL-ECBF assist system successfully boosted fuel economy and driveability while ensuring safety, even when considering distracted drivers that would cause collisions without the driver assist system in the loop.

In future works, we intend to use randomized traffic data and simulation for training and evaluation of the proposed RL-ECBF agent. Furthermore, it is necessary to consider uncertainties that are inherent in the model-based ECBF projection approach outlined here.



\bibliography{main}             
                                                   







\end{document}